\documentclass[11pt,a4paper]{article}
\usepackage[hyperref]{acl2019}
\usepackage{times}
\usepackage{latexsym}
\usepackage{titlesec}
\usepackage{chngpage}

\usepackage{url}
\usepackage{amsmath}
\usepackage{graphicx}
\usepackage{makecell}
\DeclareMathOperator*{\argmax}{argmax}
\usepackage{url}

\usepackage{microtype}

\aclfinalcopy 

\title{Multi Sense Embeddings from Topic Models}

\author{Shobhit Jain \\
  Amazon Web Services \\
  {\tt jainshob@amazon.com} \\\And
  Sravan Babu Bodapati \\
  Amazon Web Services \\
  {\tt sravanb@amazon.com} \\\AND
  Ramesh Nallapati \\
  Amazon Web Services \\
  {\tt rnallapa@amazon.com} \\\And
  Anima Anandkumar \\
 Amazon Web Services \\
 {\tt anima@amazon.com} \\}

\date{}

\begin{document}
\maketitle
\begin{abstract}
  Distributed word embeddings have yielded state-of-the-art performance in many NLP tasks, mainly due to their success in capturing useful semantic information. These representations assign only a single vector to each word whereas a large number of words are polysemous (i.e., have multiple meanings). In this work, we approach this critical problem in lexical semantics, namely that of representing various senses of polysemous words in vector spaces. We propose a topic modeling based skip-gram approach for learning multi-prototype word embeddings. We also introduce a method to prune the embeddings determined by the probabilistic representation of the word in each topic. We use our embeddings to show that they can capture the context and word similarity strongly and outperform various state-of-the-art implementations.
  
\end{abstract}

\section{Introduction}
Representing words as dense, low dimensional embeddings ~\cite{mikolov:2013a, mikolov:2013b, pennington2014glove} allow the representations to capture useful syntactic \& semantic information making them useful in downstream Natural Language Processing tasks. However, these embedding models ignore the lexical ambiguity among different meanings of the same word. They assign only a single vector representative of all the different meanings of a word. In this work, we attempt to address this problem by capturing the multiple senses of a word using the global semantics of the document in which the word appears. \citet{li2015multi} indicated that such sense specific vectors improve the performance of applications related to semantic understanding, such as Named Entity Recognition, Part-Of-Speech tagging.

In this work, we first train a topic model on our corpus to extract the topic distribution for each document. We treat these extracted topics as a heuristic to model word senses. We hypothesize that these word senses correlate quite well with the human notion of word senses, and validate it through our rigorous experiments as we demonstrate in our results section. We then use this topic distribution to train sense-specific word embeddings for each sense. We train these embeddings by weighing the learning procedure in proportion to the corresponding topic representation for each document. However, a word need not strongly correlate with each of these extracted senses. To address it, we propose a variant of this model which restricts the learning to only those embeddings where the word has a strong correlation with the topic extracted, i.e., high $p(word|topic)$.

The major contributions of our work are (i) training multi-sense word embeddings based on structured skip gram using topic models as a precursor (ii) non-parametric approach which prunes the embeddings to capture variability in the number of word senses. 

\section{Prior Work}

Recently, learning multi-sense word embedding models has been an active area of research and has gained a lot of interest. TF-IDF \citep{reisinger2010multi}, SaSA \citep{wu2015sense}, MSSG \citep{neelakantan2015}, \citet{huang2012} used cluster-based techniques to cluster the context of a word and comprehend word senses from the cluster centroids. \citet{tian2014probabilistic} proposed to use EM-based probabilistic clustering to assign word senses. \citet{li2015multi} used Chinese Restaurant Process to model the word senses. All these techniques are just local context based and thus ignore the essential correlations amongst words and phrases in a broader document-level context. In contrast, our method enriches the embeddings with the document level information, capturing word interactions in a broader document-level context.

AutoExtend \citep{rothe2015autoextend}, Sensembed \citep{iacobacci2015sensembed}, Nasari \citep{ camacho2016nasari}, Deconf \citep{pilehvar2016conflated}, \citet{chen2014unified, jauhar2015ontologically, pelevina2017making} have used multi-step approach to learn sense \& word embeddings but require an external lexical database like WordNet to achieve it. SW2V\citep{mancini2016embedding} train the embeddings in a single joint training phase. Nonetheless, all these methods assign same weight to every sense of a word, ignoring the extent to which each sense is associated with it's context.

MSWE \citep{nguyen2017mixture} trained sense and word embeddings separately, with sense specific word embeddings computed as a weighted sum of the two, where the weights are calculated using topic modeling. Similarly, \citet{liu2015learning, liu2015topical, cheng2015contextual, zhang2016improving} used skip-gram based approach to obtain separate word \& topic embeddings. \citet{lau2013unimelb} also used topic models to distinguish between different senses of a word. All these techniques express the sense-specific word representation as a function of word \& sense embeddings which essentially belongs to two different domains. Our work trains more robust compositional word embeddings formulated as a weighted sum of sense specific word embeddings, thus, taking into consideration all the different word senses while operating in the same vector space.

More recent techniques like ELMo~\citep{peters2018deep}, BERT~\citep{devlin2018bert} compute the contextual representations of a word based on the sentence in which the word appears, whereas, our method yields precomputed embeddings for each sense of a word within the same vector space.

\section{Multi Sense Embeddings Model}
\subsection{Topic Modeling}
    Mixed membership models like topic models allow us to discover “topics” that occur in a collection of documents. A \textit{topic} is defined as a distribution over words and consists of cluster of words that occur frequently. This formulation benefits us in inferring the probability distribution over different contexts(topics) the word can occur in. Latent Dirichlet Allocation(LDA) \cite{blei2003LDA} is a topic modeling technique that assigns multiple topics in different proportions to each document along with the probability distribution over words for each of the topics. Topic models based on Gibbs Sampling~\citep{geman1987stochastic} achieve this by computing the posterior for a word based on the topic proportion at document level coupled with how often the word appears together with other words in the topic. We use Gibbs Sampling based approach to compute the topic distribution for each document. We use the LDA implementation from MALLET \textit{topic modeling} toolkit~\citep{McCallumMALLET} for our experiments.
    
\begin{figure}[t]
  \includegraphics[width=\linewidth]{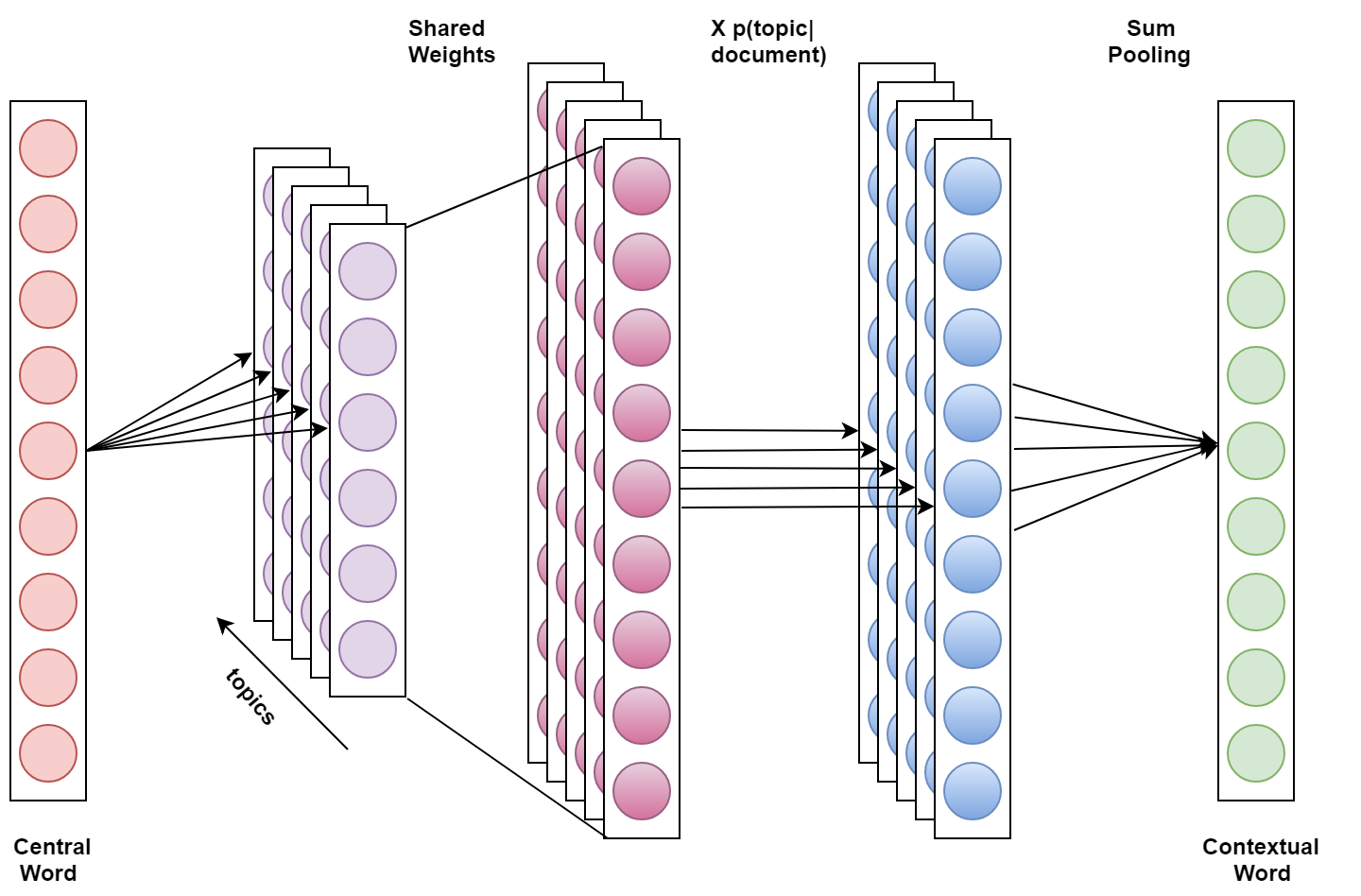}
  \caption{We feed our model a central word as input and predict the context word from it. First we learn separate word embeddings corresponding to each topic, $E_{w_{t},z_{i}}$. Each of these embeddings are then multiplied with global word embeddings, $v_{g}(w)$, weighed in proportion to the topic distribution of the document from which the words have been chosen, and summed up to predict the neighboring context word.}
  \label{fig:architecture}
\end{figure}

\subsection{Embeddings from Topic Models (ETMo)} \label{etmo}
In this section we present our baseline approach for training sense-specific word embeddings.
We formulate our approach as follows. Let $E_{w}\in R^{k \times n}$ represent the embedding matrix for word $w$, where $k$ is the number of topics(treated as number of word senses) and $n$ is the dimensionality of embeddings. We represent the embedding of word $w$ corresponding to topic $z_{i}$ as $E_{w,z_{i}}$. Let $v_{g}(w)$ be the \textit{output} vector representation for word $w$, which is shared across senses, and enforces the embeddings of different senses to be within the same vector space.

We introduce a latent variable $z$, representing the topic dimension, to model separate embedding for each topic. Inline with the skip-gram\cite{mikolov:2013a} approach, we maximize the probability of predicting the context word $w_{t+j}$, given a central word $w_{t}$ for a document $d$ as:
\begin{equation}
\begin{split}
    &p(w_{t+j}|w_{t}, d) = \\
    \sum_{i=1}^{k} &p(w_{t+j}|w_{t}, z_{i}, d) * p(z_{i}|d)
\end{split}
\end{equation}
$p(z_{i}|d)$ represents the topic distribution of the document $d$, obtained from the trained topic model. In the above equation, we reasonably make the assumption, $p(z_i | w_t, d)$ = $p(z_{i}|d)$, owing to the fact that the topic distribution is computed at the document level. Using Negative Sampling~\citep{mikolov:2013b}, we reduce the first term in the above equation as: 

\begin{equation}
\begin{split}
    p(w_{t+j}|w_{t}, z_{i})& = \\
    \sigma(E_{w_{t},z_{i}}*v_{g}(w_{t+j})) + \sum_{w\in S} \sigma(&-E_{w_{t},z_{i}}*v_{g}(w))
\end{split}
\end{equation}

Formally, given a large corpus of documents, with size $D$, having a words sequence \textit{$w_{1},w_{2}, ...,w_{N_{d}-1}, w_{N_{d}}$}, where $N_{d}$ is the number of words in document $d$, skip-window size $c$, number of topics $k$, the objective is to maximize the following log likelihood:
\begin{equation}
\begin{split}
    L &= \sum_{d=1}^{D} \sum_{t=1}^{N_{d}} \sum_{j=-c}^{c}log\ p(w_{t+j}|w_{t}, d) = \\
    \sum_{d=1}^{D} \sum_{t=1}^{N_{d}}& \sum_{j=-c}^{c}log\ \sum_{i=1}^{k} p(w_{t+j}|w_{t}, z_{i}, d) * p(z_{i}|d)
\end{split}
\end{equation}
As shown in Figure \ref{fig:architecture}, we use a neural network architecture to compute the log likelihood. We feed the central word, in its BoW representation, as input to the model and compute the probability of the context word. Refer to the figure for detailed explanation.\\

During inference, we first compute the topic distribution for the given document, $p(z_{i}|d)$, using our pre-trained topic model. Finally, for a document $d$ and for each word $w$, we infer the word embedding as:
\begin{equation}
    v_{w,d} = \sum_{i=1}^{k}  p(z_{i}|d) * E_{w,z_{i}}
\end{equation}

 \subsection{ETMo + Non-parametric}
 In this section, we substantiate the flaws in our baseline approach and present our non-parametric method to learn the embeddings.\\
  Our previous approach assigns an embedding to every word corresponding to each topic. As one can see, this method would undesirably accumulate a fair amount of noisy updates to those word embeddings that have minimal representation in a topic. Hence, we extend our model by exploiting the information from topic models to learn only those embeddings where the word has a strong correlation with the topic. \\
  In particular, we train only those embedding $E_{w_{t},z_{i}}$ such that $p(w_{t}|z_{i}) > p_{thres}$, where $p_{thres}$ is chosen empirically, which we will explain later. For the words where none of the senses satisfy the above condition (might be the case for some monosemous words), we chose the embedding $E_{w_{t}, x}$ to be trained, such that $x = \argmax_{z_{i}} p(w_{t}|z_{i})$.

\section{Experimental Setup} \label{experiments}

\begin{table}[t]
\fontsize{10pt}{13pt}\selectfont
\centering
\begin{adjustwidth}{-.05in}{-.05in}
\begin{tabular}{l r r}
    \hline
    \textbf{Model} & $avgSim$ & $globalSim$\\
    \hline
    GloVe & - & 63.2\\
    \citep{pennington2014glove} & & \\
    \citet{huang2012} & 64.2 & 71.3\\
    csmRNN & - & 64.58 \\
     \citep{luong2013better} & & \\
    GC-SINGLE & 62.3 & - \\
     \citep{jauhar2015ontologically} & & \\
     NP-MSSG & 69.1 & 68.6 \\
     \citep{neelakantan2015} & & \\
     MSWE-I & - & \textbf{72.40} \\
     \citep{nguyen2017mixture} & & \\
    Gensense  & 54.0 & - \\
    \citep{lee2018gensense} & & \\
    \hline
    ETMo (Ours)  & 68.5 &  68.2\\
    ETMo + NP (Ours)  & \textbf{69.3} &  69.1\\
    \hline
\end{tabular}
\end{adjustwidth}
\caption{\label{ws} Spearman's correlation $\rho \times 100$ on WS-353}
\end{table}

 We use the English Wikipedia corpus dump ~\citep{wikipedia} for training both, topic models and embedding models. Though many previous research works have used a larger training corpus, but for a fair comparison, we only compare our results with those works which have used the same corpus. We could also improve obtained results by using a larger training corpus, but this is not central point of our paper. The main aim of our work is to compute sense specific embeddings for a word using topic models and demonstrate the strength of our model empirically. 
 
 The raw dataset consists of nearly 3.2 million documents and 1 billion tokens. Training topic models on such a large and diverse corpus helps in obtaining clearly demarcated senses for each topic. 
 
 To tune the hyper parameters of our neural network model, we sample 20\% of our corpus as validation data and chose those parameters that give the lowest validation loss. Later, we use these parameter values for training on the entire corpus. For all our experiments, we use a skip-window of size 2, 8 negative samples, embeddings of dimensionality 200, and fix the number of topics to 10. A detailed analysis on how we chose the number of topics, using perplexity score, can be found later in the analysis section. We initialize the embeddings using pre-trained GloVe embeddings to ensure all our target embeddings are in the same vector space. We choose the value for $p_{thres}$ as 1e-4 and give an analysis on how we chose the parameter value in the results section.

\begin{table}[t]
\centering
\fontsize{10pt}{12pt}\selectfont
\begin{tabular}{l r r }
    \hline
    \textbf{Model} & $avgSim$ & $avgSimC$\\
    \hline
    TF-IDF & 60.4 & - \\
    \citet{huang2012} & 62.8 & 65.7\\
    \citet{tian2014probabilistic} & - & 65.4 \\
   \citet{chen2014unified} & 66.2 & 68.9 \\
   \citet{cheng2015contextual} & - & 65.9 \\
   GC-MULTI & - & 65.9 \\
    \citep{jauhar2015ontologically} & & \\
   SENSEMBED & 62.4 & - \\
    \citep{iacobacci2015sensembed} & & \\
   SaSA & - & 66.4 \\
   \citep{wu2015sense} & & \\
   TWE-I \citep{liu2015topical} & - & 68.1 \\
	NP-MSSG & 67.2 & \textbf{69.3} \\
	 \citep{neelakantan2015} & & \\
   SG+Greedy & - & 69.1 \\
    \citep{li2015multi} & & \\
    MSWE & 66.7 & 66.6 \\
    \citep{nguyen2017mixture} & & \\
     \hline
    ETMo (Ours) & 65.4 & 65.8\\
    ETMo + NP (Ours) & \textbf{67.5} & 69.1 \\
   \hline
\end{tabular}
\caption{\label{results}Spearman's correlation $\rho \times 100$ on SCWS}
\end{table}

\begin{table}[t]
\centering
\begin{tabular}{ l r }
    \hline
    Model & Accuracy(\%) \\
    \hline
    Word2Vec & 67 \\
    \citet{huang2012} & 12 \\
    \citet{neelakantan2015} & 64 \\
    ETMo (Ours) & \textbf{67} \\
    ETMo + NP (Ours) & 66 \\
    \hline
\end{tabular}
\caption{\label{analogy} Results on Word Analogy task}
\end{table}

\section{Results} \label{section5}
We evaluate our model on two tasks, namely, word similarity and word analogy. For word similarity evaluation, we evaluate our embeddings on standard word similarity benchmark datasets including WS-353 ~\citep{finkelstein2001placing} \& SCWS-2003 ~\citep{huang2012}. WS-353 includes 353 pairs of words and a human judgment score of the similarity measure between the two words. Similarly, SCWS-2003 consists of 2003 pairs of words, but, given with a context.

We note that our embeddings can capture only those senses that are represented by the extracted topics, and due to the restricted number of topics extracted, they might not be able to capture all the senses for a word. However, at a specific number of topics, our model is effective in capturing various senses of words in standard word similarity datasets. We demonstrate this effect qualitatively and quantitatively in this section.

For each of the datasets, we report the Spearman correlation between the human judgment score and model's similarity score computed between two words $w$ and $w'$. We follow \citet{reisinger2010multi} to compute the following similarity measures. For a pair of words $w$ and $w'$ and given their respective contexts $c$ and $c'$, we represent the cosine distance between the embeddings $E_{w,i}$ and $E_{w',j}$ as $d(E_{w,i}, E_{w',j})$.
\begin{equation}
\begin{split}
    globalSim& = d(v_{g}(w), Ç) \\
    avgSim =&  \frac{1}{N_{1}*N_{2}}\sum_{i=1}^{N_{1}} \sum_{j=1}^{N_{2}} d(E_{w,i}, E_{w',j}) \\
	 avgSimC& =  \sum_{i=1}^{N_{1}} \sum_{j=1}^{N_{2}} p(z_{i}|c)*p(z_{j}|c')*\\     &    d(E_{w,i}, E_{w',j})
\end{split}
\end{equation}
$N_{1}$ and $N_{2}$ are chosen such that they satisfy $p(w_{t}|z_{i}) > p_{thres}$. $v_{g}(w)$ represents the \textit{output} vector for word $w$, as mentioned in section \ref{etmo}. We infer the probabilities, $p(z_{i}|c)$ \& $p(z_{j}|c')$ using our pre-trained topic model. 


In contrast to our model, methods such as ELMo, BERT requires a document context to compute an embedding, which makes it unfair to compare on avgSim metric since it doesn't take any context into account. Additionally, ELMo gives a set of 3 different embeddings making it unclear to compare on the avgSimC metric as well. 

\begin{table*}
\fontsize{10pt}{12pt}\selectfont
\centering
\begin{tabular}{|l|l|l|}
\hline
  \textbf{Word}  & \textbf{Topic \#} & \textbf{Nearest Neighbors}\\
  \hline
   & Glove & playing, played, plays, game, players, player, match, matches, games\\
   play & 2 & played. performance, musical, performed, plays, stage, release, song, work, time \\
   & 7 & season, players, played, one, game, first, football, teams, last, year, clubs \\
  \hline
    & Glove & band, punk, pop, bands, album, rocks, music, indie, singer, albums, songs, rockers \\
    rock & 2 & metal, pop, punk, members, jazz, alternative, indie, folk, band, hard, recorded, blues\\
   & 6 & island, point, valley, hill, large, creek, granite, railroad, river, lake\\
   \hline
    & Glove & banks, banking, central, credit, bankers, financial, investment, lending, citibank\\
  bank & 6 & river, tributary, flows, valley, side, banks, mississippi, south, north, mouth, branch\\
   & 8 &  company, established, central, first, group, one, investment, organisation, development\\
 \hline
    & Glove & plants, factory, facility, flowering, produce, reactor, factories, production\\
   plant & 1 & plants, bird, genus, frog, rodent, flowering, fish, species, tree, endemic, asteraceae\\
   & 5 & design, plants, modern, power, process, technology, standard, substance, production\\
 \hline
 & Glove & wars, conflict, battle, civil, military, invasion, forces, fought, fighting, wartime\\
   war & 4 & combat, first, world, army, served, american, battle, civil, outbreak, forces\\
   & 7 & series, championship, cup, fifa, champion, chess, records, wrestling, championships\\
 \hline
 & Glove & cable, channel, television, broadcast, internet, stations, programming, radio\\
   network & 2 & series, program, shows, bbc, broadcast, station, channel, aired, nbc, radio, episode\\
   & 5 & data, information, computer, system, applications, technology, control, standard, design\\
   & 6 & light, station, car, stations, railway, commuter, lines, rail, trains, commute\\
 \hline
\end{tabular}
\caption{Nearest neighbours of some polysemous words for Glove, and for each sense identified by our algorithm, based on the cosine similarility. We take only those senses corresponding to topics where $p(w_{t}|j)>p_{thres}$.}
\label{qualitative}
\end{table*}

\subsection{Quantitative Results}
We present the results of our approach in Tables \ref{ws} and \ref{results}. A higher Spearman's correlation translates to a better model. 

As can be seen in Table \ref{ws}, our non-parametric approach clearly outperforms other multi-sense embeddings models using the \textit{avgSim} metric on WS-353. Further, though our method focusses on sense specific embeddings and not on the global word embeddings, for the purpose of completeness, we also report our results on the \textit{globalSim} metric. Using \textit{globalSim}, expectedly we obtain slightly lower results since \textit{globalSim} is more suited for global word embeddings. 

In Table \ref{results}, we compare our models on the SCWS dataset. Using \textit{avgSim} metric, our model obtain state-of-the-art results, outperforming other embeddings model. Using the \textit{avgSimC} metric, we produce competitive results and perform better than most of the models, including \citet{nguyen2017mixture} which also uses topic models. 

These superior results indicate the usefulness of our method to accurately capture word representations that can take into account different word senses. Additionally, our non-parametric approach consistently outperforms our baseline ETMo approach, validating our hypothesis to threshold the topics.

We also evaluate our model on the word analogy task \citep{mikolov:2013a}. ~\footnote{ The word analogy task aims to answer the question of the form: \textit{a is to b as c is to ?}. To answer the question, we compute the word vector nearest to `$v_{g}(b)-v_{g}(a)+v_{g}(c)$', where $v_{g}(w)$ represents the \textit{output} vector for word $w$, as mentioned in section \ref{etmo}.} Our answer is correct if this word matches the correct word given in the dataset. As can seen in Table \ref{analogy}, our ETMo approach obtains similar results as the baseline word2vec model, and we beat other implementations.

\subsection{Qualitative Comparison}
We show a qualitative comparison of some polysemous words in Table \ref{qualitative}, with the nearest neighbors of words in the table, for Glove embeddings and the embeddings trained from our model.
For each of the words in Table \ref{qualitative}, we can clearly see that the different senses of words are being effectively captured by our model whereas Glove embeddings could only capture most frequently used meaning for the word. Moreover, each of these senses can be easily correlated with the topic that these embeddings correspond to which can be seen from Table \ref{topic_keys}. Consider the word \textit{Play}. The first sense for \textit{play} corresponds to \textit{Music} (topic 2). The second embedding corresponds to \textit{Sports} (topic 7). 

An interesting qualitative result is shown for the word \textit{Network}. The nearest neighbors to Glove embeddings  show that they are only able to capture one meaning which is in the subject of \textit{Television Network}. However, our model is able to capture 3 different meanings for the word quite powerfully. The first one, which corresponds to topic 2, occurs in the context of \textit{Television Network} which is the sense Glove was able to capture. The second sense, which corresponds to topic 5, occurs in the context of \textit{Computer Networks}. The third sense, which corresponds to topic 6, remarkably relates to the context \textit{Geography}. 

\begin{table*}
\fontsize{10pt}{12pt}\selectfont
\centering
\begin{tabular}{|l|c|l|}
    \hline
    \textbf{TOPIC \#} & \textbf{TOPIC KEYS} & \textbf{TOPIC NAME}\\
    \hline
    1 & \shortstack{species south india island north found small indian region  family \\ district water large east long spanish central village west area} & Agriculture\\
    \hline
    2 & \shortstack{film music album released band series show song time television \\ single songs live rock records video release appeared episode films} & Music/Television \\
    \hline
    3 & \shortstack{party government state states united president law member general\\ court house election served political elected national born council} & Politics\\
    \hline
    4 & \shortstack{war air army force british battle service aircraft japanese forces \\ world military time ship fire navy command attack september car} & Military\\
    \hline
    5 & \shortstack{system formula number time called form data systems process high \\ energy type common set space based power similar standard} & Technology\\
    \hline
    6 & \shortstack{city age county area population town located years north river \\ south west station park line road district village income living} & Geography\\
    \hline
    7 & \shortstack{team season game league played club football games world year \\ career player born time final cup play championship national} & Sports\\
    \hline
    8 & \shortstack{school university college company students education, public program \\ business national research development services million service} & Education\\
    \hline
    9 & \shortstack{church book work life published century time english works \\ art people world books language great written god early death called} & Religion\\
    \hline
 10 & \shortstack{french german france war king germany century russian part \\ italian son chinese empire soviet republic born died emperor paris} & History\\
 \hline
\end{tabular}
\caption{\label{topic_keys} The top words for each topics according to topic modeling}
\end{table*}


\subsection{Number of Topics Analysis}
In this section, we perform a study on choosing the right number of topics(k) in Table \ref{topics}. Here, topic uniqueness refers to the proportion of unique words in a topic, computed over the top words in the vocabulary. Higher the topic uniqueness score, more distinct are the obtained topics. We compute the Spearman correlation on the \textit{avgSim} metric using the word pairs from RG-65~\citep{rubenstein1965contextual}. With k = 10, we obtained a topic uniqueness of $32.23$, which dropped to $27.12$ for k=20 topics. Thus increasing the number of topics increases overlap which harms our model as the topic weight gets divided while training the embeddings. This effect can be clearly seen in the correlation coefficient which drops from $68.5$ to $66.9$ for 10 \& 20 topics respectively. Using k=5 improved the topic uniqueness score to $34.05$, but the perplexity score~\citep{blei2003LDA} reduced, indicating that the topic model requires more degrees of freedom to fit the corpus. We also observed not very distinct topics at k=5  (i.e. a topic could be mixture of sports and history), resulting in reduced correlation coefficient of $67.1$.

\begin{table}
\centering
\fontsize{10pt}{12pt}\selectfont
\begin{tabular}{|l|c|c|r| p{2cm}p{2cm}p{1.5cm}p{1.5cm}}
    \hline
    \textbf{\# of topics} &  \textbf{uniqueness} &  \textbf{perplexity} & $\rho \times 100$\\
    \hline
    5  & 34.05 & 9.88 & 67.1 \\
    \hline
    10 & 32.23 & 9.78 & 68.5\\
    \hline
    15 & 29.57 & 9.70 & 67.8\\
    \hline
    20 & 27.12 & 9.65 & 66.9\\
    \hline
\end{tabular}
\caption{\label{topics} effect of number of topics on Spearman correlation on 50 word pairs from WordSim-353}
\end{table}

\subsection{Threshold Parameter Analysis}
In this section, we study the effect of $p_{thres}$ on the model performance. We tune its value by comparing the Spearman correlation on the \textit{avgSim} metric using the word pairs from RG-65~\citep{rubenstein1965contextual}. However, we hypothesize that the threshold parameter depends only on the output of topic modeling, particularly $p(word|topic)$, and thus is independent of the this chosen subset, as can be seen in the results on other datasets. In Table \ref{pthres}, we can see that the optimal value for $p_{thres}$ is 1e-4 for the non-parametric model at which it can strongly differentiate between the different senses for \textit{network}. A higher threshold value of 1e-3 captures a fewer number of senses. A lower threshold value of 1e-5 allows training of more than the actual number of true senses leading to noisy updates, thus becoming ineffective in capturing any sense. The corresponding lower correlation coefficients in Table \ref{pthres} confirm these effects quantitatively. 

\begin{table}[t]
\centering
\begin{tabular}{l r l}
    \hline
    \textbf{$p_{thres}$} & $\rho \times 100$ & senses captured for \textit{network}\\
    \hline
    1e-3  & 68.3 & television, IT\\
    1e-4 & 69.1 & television, IT, transportation\\
    1e-5  & 68.4 & mixed senses\\
    \hline
\end{tabular}
\caption{\label{pthres} Spearman's correlation $\rho \times 100$ on 50 word pairs from WS-353 and the word senses captured for \textit{network}. The word senses are adjudged qualitatively.}
\end{table}

\section{Conclusion \& Future Work}
In this work, we presented our approach to learn word embeddings to capture the different senses of a word. Unlike previous sense-based models, our model exploits knowledge from topic modeling to induce mixture weights in structured skip-gram approach, for learning sense specific representations. We extend this model further by pruning the embeddings conditioned on the number of word senses. Finally, we showed our model achieves state-of-the-art results on word similarity tasks, and demonstrated the strength of our model in capturing multiple word senses qualitatively. Future work should aim towards using these embeddings for downstream tasks.



\newpage

\bibliography{acl2019}
\bibliographystyle{acl_natbib}

\appendix

\end{document}